\documentclass[runningheads]{llncs}
\usepackage{graphicx}
\usepackage{comment}
\usepackage{amsmath,amssymb, amsfonts} 
\usepackage{color}
\usepackage{float} 
\usepackage{siunitx}
\usepackage{textcomp}
\usepackage{cite}
\begin{document}
\pagestyle{headings}
\mainmatter
\def\ECCVSubNumber{6696}  

\title{Scan2Plan: Efficient Floorplan Generation from 3D Scans of Indoor Scenes }

\titlerunning{Scan2Plan}
\author{Ameya Phalak \and
Vijay Badrinarayanan \and
Andrew Rabinovich}
\authorrunning{Phalak et al.}
\institute{Magic Leap, Inc.\\
Sunnyvale, CA 94089, USA \\
\email{\{aphalak, vbadrinarayanan, arabinovich\}@magicleap.com}} 

\maketitle


\begin{abstract}

We introduce Scan2Plan, a novel approach for accurate estimation of a floorplan from a 3D scan of the structural elements of indoor environments. The proposed method incorporates a two-stage approach where the initial stage clusters an unordered point cloud representation of the scene into room instances and wall instances using a deep neural network based voting approach. The subsequent stage estimates a closed perimeter, parameterized by a simple polygon, for each individual room by finding the shortest path along the predicted room and wall keypoints. The final floorplan is simply an assembly of all such room perimeters in the global co-ordinate system. The Scan2Plan pipeline produces accurate floorplans for complex layouts, is highly parallelizable and extremely efficient compared to existing methods. The voting module is trained only on synthetic data and evaluated on publicly available Structured3D and BKE datasets to demonstrate excellent qualitative and quantitative results outperforming state-of-the-art techniques.
\end{abstract}
\section{Introduction}
\label{sect:introduction}
A detailed understanding of the semantic components that constitute an indoor environment is gradually growing into an issue of increasing importance. Such insights, which fall under a broad topic that is popularly known as Scene Understanding, are obtained in various ways - for example semantic segmentation of 2D/3D data of indoor environments\cite{he2017mask, zhao2017pyramid}, object detection/recognition\cite{redmon2016you, ren2015faster, liu2016ssd}, CAD scan replacement of furniture\cite{avetisyan2019scan2cad}, floorplan estimation \cite{liu2018floornet, cjc2019floorsp} among others. 

In this work, we focus on the task of efficiently generating an accurate floorplan of an indoor scene to aid such Scene Understanding. The capability of generating a floorplan from a 3D scan has far reaching implications in multiple academic and commercial domains. The housing industry, architecture design and interior design are being pervaded by technology more than before, and automated tools such as Scan2Plan can greatly increase the efficiency and the spectrum of design possibilities for such industries. Similarly, a smarter understanding of the environment is absolutely essential for Augmented and Virtual reality(AR/VR) devices to provide a contextual, richer, more interactive experience for consumers.    

Floorplan estimation is a niche task as compared to commonly seen deep learning problems such as semantic segmentation and object recognition. Additionally, the logistical difficulties associated with capturing real-world indoor environments has naturally reflected in a shortage of datasets of indoor environments with annotated floorplans and an accompanying 3D point cloud/mesh representation. Moreover, it is extremely unlikely that such a dataset of single origin is capable of possessing samples with a large number and variety of different layout types that are necessary to train a deep network capable of performing well in the wild. Therefore, we propose a procedurally generated fully synthetic dataset to overcome such barriers. 

We describe an approach to extract a floorplan of an indoor environment with single or multiple rooms from a 3D scan of its structural elements such as walls, doors and windows. The networks in our floorplan extraction pipeline, trained fully on synthetic data, are capable of estimating an unconstrained layout with no restrictions on the shape and number of rooms within the bounds of reason. We assume that state-of-the-art techniques such as MinkowskiNet\cite{choy20194d}, ScanComplete\cite{dai2018scancomplete} for 3D scans and/or MaskRCNN\cite{He_2017_ICCV}, PSPNET\cite{zhao2017pyramid} for 2D segmentation(if accompanying RGB/Grayscale images are available) should be able to create a scan of only the structural elements which Scan2Plan can then process to generate a floorplan.

\begin{figure}
\centering
\includegraphics[width=\textwidth]{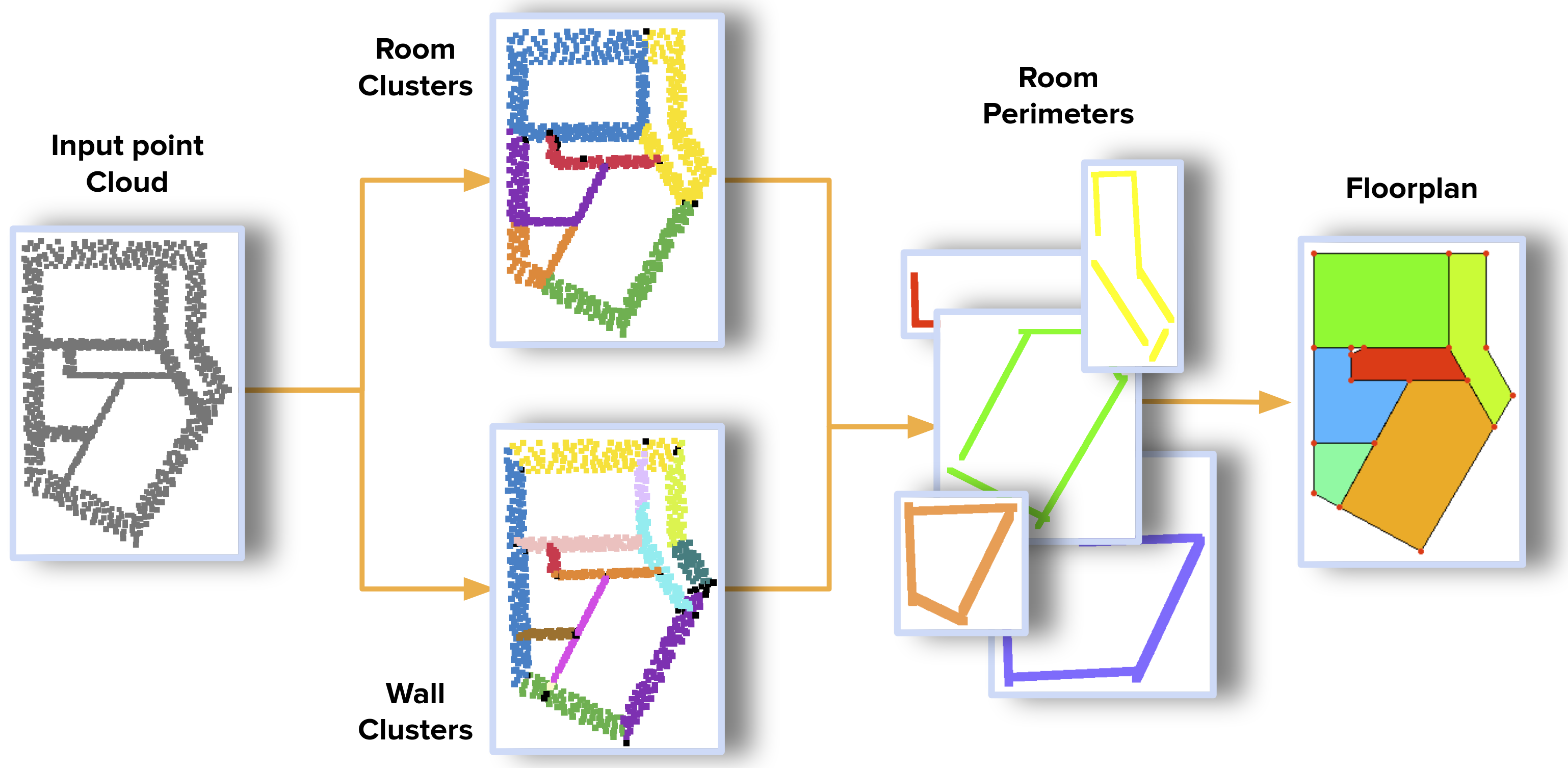}
\caption{\textbf{Scan2Plan} is a robust and efficient method that extracts a detailed floorplan from a 3D scan of an indoor environment.}
\label{fig:teaser}
\end{figure}

In related methods such as FloorSP\cite{cjc2019floorsp} which use a global graph optimization technique, the complexity of the problem does not scale well with the size of the input in terms of number of rooms and the number of walls per room, which, in turn is proportional to the size of the graph to be optimized. In contrast Scan2Plan combats the scaling issue by breaking down the global problem into multiple smaller local problems which can be processed efficiently in a parallel fashion.

Our method broadly follows a two step approach. Firstly, a deep network based on a Pointnet++\cite{qi2017pointnet++} backbone generates cluster labels for the vertices of a perceptually uniform subset of the input point cloud in order to identify the rooms/areas that comprise the indoor scene. Simultaneously, the network also generates another set of cluster labels to identify  each wall in the given scene. Secondly, we jointly utilize the room and wall labels to create a DeepPerimeter\cite{phalak2019deepperimeter} type of shape for every room, expressed by a simple polygon. The final floorplan is simply a collection of all the estimated room perimeters put together in the global coordinate system of the original point cloud. An illustration of the end-to-end pipeline can be seen in Fig. \ref{fig:teaser}

In summary, we make the following key contributions through this work:

(1) We propose a novel technique to generate a detailed floorplan from a 3D scan of an indoor scene. Our method does not impose any constraints on the number of rooms, their configurations or their shapes.

(2) We demonstrate that the problem of floorplan estimation can be solved by training deep networks on a purely synthetic dataset which can be generated efficiently and is also highly configurable in all aspects.

(3) The proposed solution is highly parallelizable in multiple stages and outperforms the current state-of-the-art in terms of run-time efficiency as well as accuracy.
\section{Related Work}

\textbf{Feature Extraction on Unordered Point Clouds}
With the increasing momentum of the augmented and virtual reality industry and the autonomous vehicle industry, availability of 2.5D and 3D data from various sources such as smartphones, consumer-grade depth sensors, LiDAR systems has seen a significant boost in recent times. Subsequently, the need to perform complex learning tasks on such data has also seen a rise in demand. Architectures such as PointNet\cite{qi2017pointnet} and its successor PointNet++\cite{qi2017pointnet++} operate directly upon the point vectors whereas others partition the input space into a structured grid\cite{zhou2018voxelnet, wang2017cnn, tatarchenko2017octree, li2018pointcnn} and quantize the input so it is possible to run 3D convolutions on the derived grid. More recent approaches such as MinkowskiNet\cite{choy20194d} perform sparse convolutions on a 3D point set to achieve impressive results for object detection, while graph-based approaches\cite{wang2019dynamic, qi20173d, shen2018mining, phan2018dgcnn} parameterize the input data points into a set of nodes and edges and exploit the graph connectivity and structure to extract features.

\textbf{Clustering}
The problem of clustering can be broadly defined as a label assignment task where data points with similar features are to be assigned the same label. Recently, deep neural networks have been utilized to perform this task in a supervised or semi-supervised\cite{qi2019deep, shukla2018semi} and unsupervised\cite{phalak2019deepperimeter, caron2018deep, xie2016unsupervised} setting. Similar to our work, some prior research also explores a voting mechanism\cite{qi2019deep, dimitriadou2001voting, ayad2007cumulative, iqbal2012semi} for clustering.


\textbf{Floorplan Estimation}
Prior research on floorplan estimation has a large variation in the parameterization practices of output floorplans, since there is not an universal and standardized way of expressing such a representation. Similarly, an indoor environment can be captured in a variety of ways depending on the availability of sensor suites at hand and also the dimension and the desired resolution of the capture. Traditional methods like \cite{zhang2013estimating} use panoramic RGBD images to procedurally reconstruct a space using structure grammar, whereas \cite{murali2017indoor} uses a 3D scan to extract plane primitives and heuristically generate building information models. Certain deep learning based methods process a single image(pinhole or panoramic) to generate a cuboid-based layout for a single room\cite{zhang2019edge, hsiao2019flat2layout, Lee_2017_ICCV, dasgupta2016delay, zou2018layoutnet, kruzhilov2019double}. Typically these methods rely on visual cues such as vanishing points and wall edges to estimate a layout, often selecting the best match from a library of pre-compiled manhattan-style room shapes. DeepPerimeter\cite{phalak2019deepperimeter} is also a single room layout estimation method that generates a polygonal perimeter using a monocular video of the scene. To estimate a floorplan of an apartment/house\cite{liu2018floornet, cjc2019floorsp} process a 3D scan to generate a relatively more detailed and semi-constrained floorplan. However, the high compute requirements of their methods limit the number of scans that can be processed using consumer-grade hardware in an acceptable time, such as under one minute per scan.

\begin{figure}
\centering
\includegraphics[width=\textwidth]{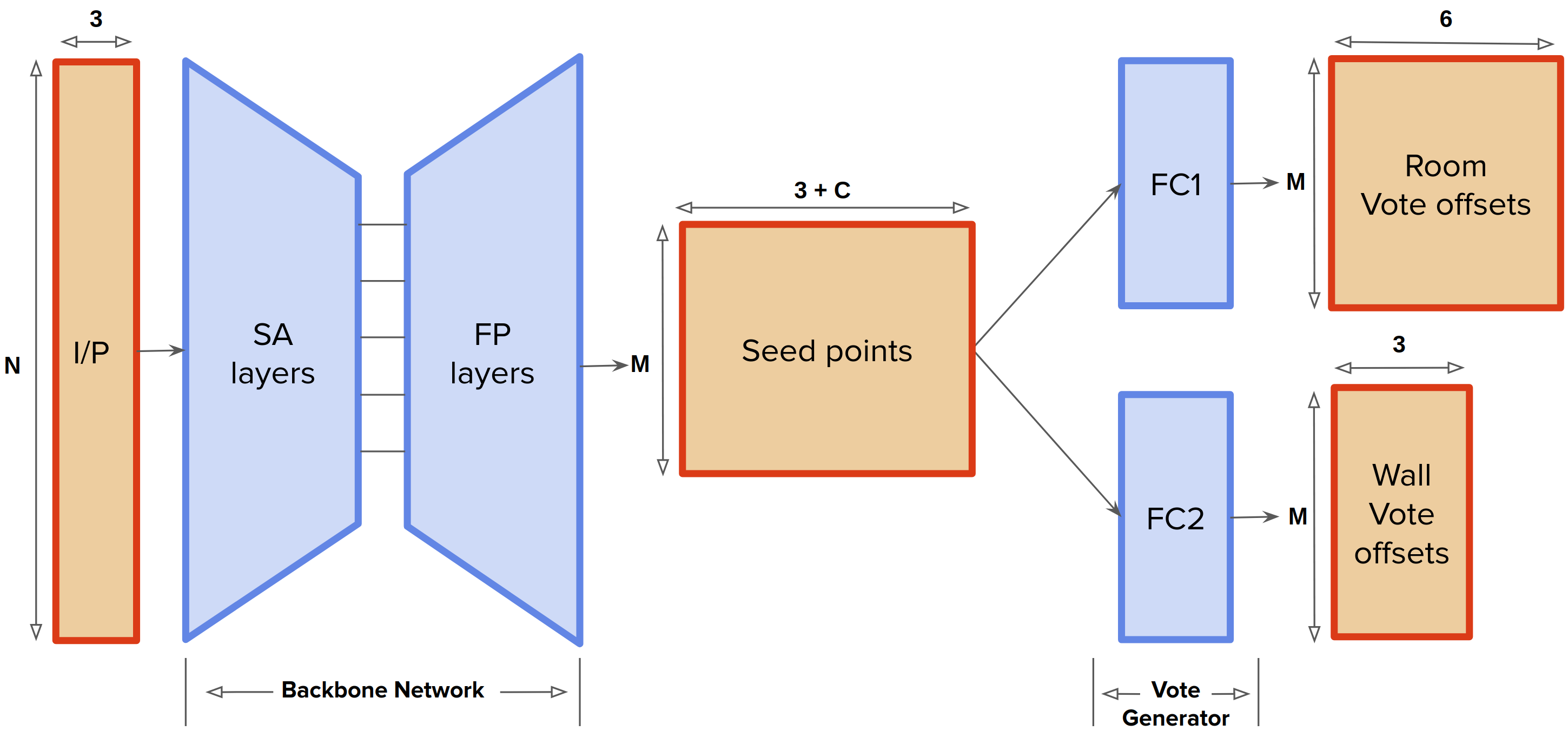}
\caption{\textbf{Architecture of the Voting Module}
The network accepts a point cloud with $N$ points in 3D Euclidean space at the input layer. A PointNet++\cite{qi2017pointnet++} backbone with set abstraction(SA) and feature propagation(FP) layers is utilized to sub-sample the input to a point cloud with $M$ 3D points, referred to as seeds and a $C$ length feature vector per seed. Vote generators FC1 and FC2, which are fully connected networks shared among all seeds, generate three 3D offset vectors per seed - two for room votes and one for wall votes.
}
\label{fig:votenet_architecture}
\end{figure}

\section{Method}
\label{sect:system}
In this section we describe in detail all the components involved in inferring a 2D floorplan from a 3D scan of an indoor scene. The key steps being - identifying room instances and wall instances for each room from the original 3D scan(Sec. \ref{subsect:votenet}) followed by estimating a closed perimeter for each room instance(Sec. \ref{subsect:perimeter}).

\subsection{Room and Wall Clustering}
\label{subsect:votenet}
We pose the problem of segregating the input 3D point cloud into its constituting rooms and walls as a mutually non-exclusive clustering of 3D data points without any prior assumptions on the number of clusters. In order to enable this step to predict an unconstrained number of clusters independent of the network architecture, we adopt a learnt voting based technique inspired from \cite{qi2019deep}.

\begin{figure}
\centering
\includegraphics[width=0.8\textwidth]{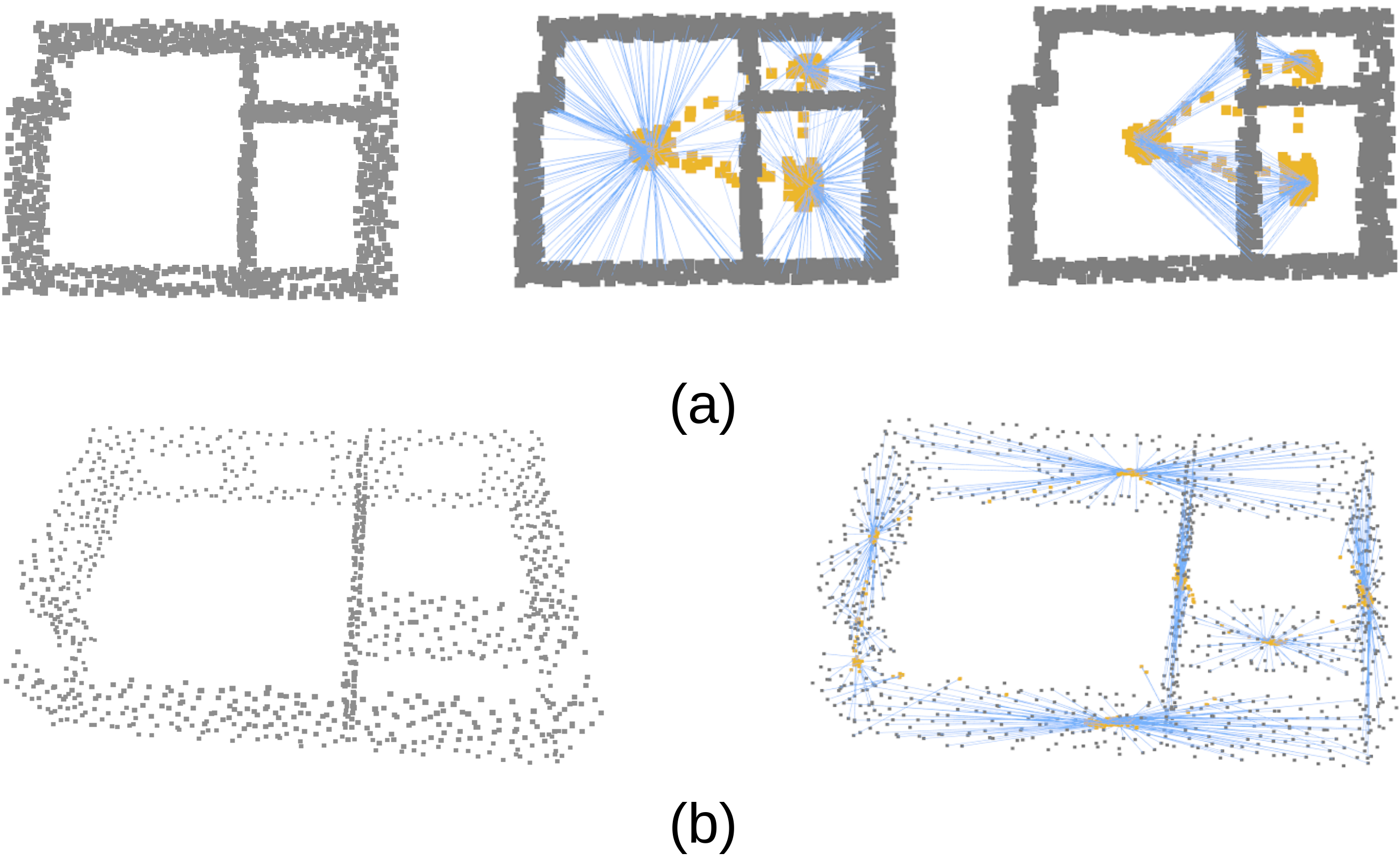}
\caption{\textbf{Voting for Room and Wall Centers} (a) From left to right - The point cloud input to the voting module, seed points $\mathcal{S}$ (gray), with vote points $\mathcal{V}$ (orange) and vote offset vectors $\mathcal{V}$ (blue) for distinct and shared room votes.  (b) From left to right - The point cloud input to the voting module, seed points $\mathcal{S}$ (gray) with vote points $\mathcal{V}$ (orange) and vote offset vectors $\mathcal{X}$ (blue) for wall votes.}
\label{fig:voting_viz}
\end{figure}

\subsubsection{Room and Wall Center Regression}
\label{subsubsect:center_regression}
The architecture of the voting module is summarized in Fig. \ref{fig:votenet_architecture}. We use a PointNet++\cite{qi2017pointnet++} backbone as our feature extractor, the input to which, is a point cloud of points $\mathcal{P}$ such that $\mathcal{P} = \{{p_j}\}_{j=1}^N$ where $p_j \in \mathbb{R}^3$. The set abstraction(down-sampling) layers and the feature propagation(up-sampling) layers in the backbone compute features at various scales to produce a sub-sampled version of the input denoted by $\mathcal{S}$, with $M$ points, $M \leq N$ having $C$ additional feature dimensions such that  $\mathcal{S} = \{{s_i}\}_{i=1}^M$ where $s_i \in \mathbb{R}^{3 + C}$. We here on refer to the set $\mathcal{S}$ as seed points, each member of which casts votes $\mathcal{V}$ such that $ \mathcal{V} = \{{v_i^q}\}_{i=1}^M$ where $v_i \in \mathbb{R}^3$ and $q \in \{R_0, R_1, W\}$. Each seed point $s_i$ thus casts 3 votes, where $q=R_0$ or $q=R_1$ implies that the vote denotes the center of room that $s_i$ belongs to, and $q=W$ implies that the vote denotes the center of the wall that $s_i$ belongs to. 

In the case where a seed point $s_i$ belongs to a single room, $v_i^{R_0}$ and $v_i^{R_1}$ are identical whereas in the case of wall points shared among two rooms, they are distinct. For votes cast to determine the wall centers $v_i^{W}$, we assume that each point can only belong to a unique wall and find this assumption to hold sufficiently in practice.

In order to generate a vote $v_i^q$ from each seed $s_i$, we use multiple vote generators, which are fully connected layers followed by Batch-Norm and ReLU layers as described in \cite{qi2019deep}. A vote generator extracts vote offsets $\mathcal{X}$ from every seed point in $\mathcal{S}$ such that $\mathcal{X} = \{{x_i^q}\}_{i=1}^M$ where $x_i \in \mathbb{R}^3$. A vote offset is nothing but an offset vector from a seed point to its vote such that $v_i^q = s_i  + x_i^q$. In practice, we use two vote generators, one for room votes where $q \in \{R_0, R_1\}$ and the other for wall votes where $q = W$. The parameters of any particular vote generator are shared among all the $M$ seed points and hence each $s_i$ is able to generate an $x_i^q$ independent of any other seed points. 

We train the network using a loss $\mathcal{L}$ given by:

\begin{equation}
\mathcal{L} = \mathcal{L}_{room} + \alpha \cdot \mathcal{L}_{wall}\,,
\end{equation}

\text{where}

\begin{equation}
\mathcal{L}_{room} = \frac{1}{M} \sum_{i}^M \ell(e_i^{q \in \{R_0, R_1\}})
\end{equation}

\text{and}

\begin{equation}
\mathcal{L}_{wall} = \frac{1}{M} \sum_{i}^M \ell(e_i^{q=W})\,,
\end{equation}

\text{such that $\ell$ is a $smooth$-$L1$ loss defined as}

\begin{equation}
\ell(a) =
        \begin{cases}
        0.5 \times (a)^2 & \text{if } |a| < 1, \\
        |a| - 0.5 & \text{otherwise}
        \end{cases}
\end{equation}

\text{and}

\begin{equation}
\label{eq:eiq}
e_i^q = 
        \begin{cases}
        \min(
        g_i^{R_0} - x_i^{R_0} +  g_i^{R_1} - x_i^{R_1},  
        g_i^{R_0} - x_i^{R_1} + g_i^{R_1} - x_i^{R_0}) & \text{if } q \in \{R_0, R_1\}, \\
            g_i^{W} - x_i^{W} & \text{otherwise},
        \end{cases}
\end{equation}
where $g_i^q$ is the ground truth offset vector corresponding to the predicted offset vector $x_i^q$ and $\alpha$ is a constant to balance the loss magnitudes.

In the case of $\mathcal{L}_{room}$ where we have two unordered pairs of ground truths and predictions, we choose a pairwise difference that results in the lowest error in Eq. \ref{eq:eiq}. This allows us the to optimize the network parameters without enforcing an artificial ordering on the room votes.
A visualization of the wall and room votes for an example from the Structured3D\cite{zheng2019structured3d} dataset is displayed in Fig. \ref{fig:voting_viz}.

\begin{figure}
\centering
\includegraphics[width=0.75\textwidth]{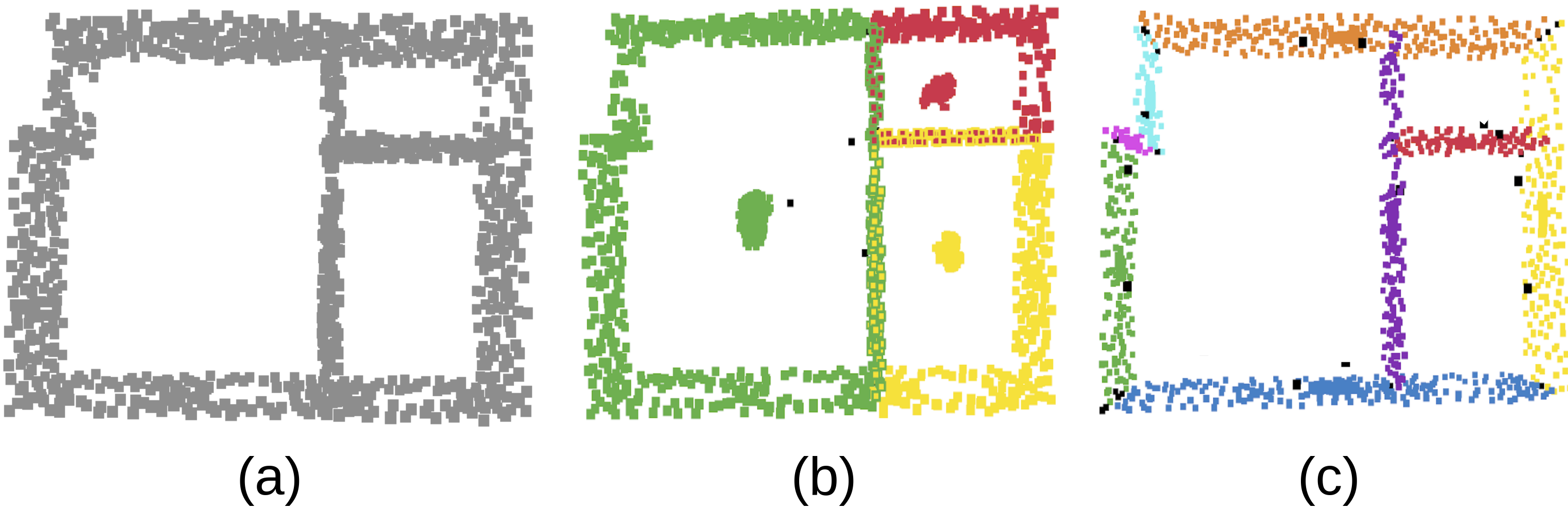}
\caption{Vote and Seed Clustering (a) - The point cloud input to the voting module. (b) and (c) - Cluster labels backtracked from the votes $\mathcal{V}$ to the seed points $\mathcal{S}$ for room and wall labels respectively.}
\label{fig:clustering_viz}
\end{figure}

\subsubsection{Vote Clustering}
\label{subsubsect:clustering}
Given the set of predicted centers $v_i^q$, we now intend to generate corresponding cluster labels for the seed points to obtain room and wall instances. To perform this operation we concatenate $v_i^{R_0}$ and $v_i^{R_1}$ to prepare clustering input for rooms, and simply use $v_i^W$ for walls. We run a spatial density based clustering algorithm - DBSCAN\cite{ester1996density, schubert2017dbscan} with eps=$\epsilon$ on the room and wall votes separately and assign a cluster label $l_i^{q'}$ to each vote $v_i^{q'}$, such that $q' \in \{ R, W \}$ where $q' = R$ implies the label is a cluster assignment for rooms, whereas $q'= W$ implies that the label is a cluster assignment for walls. Using DBSCAN allows us to cluster the votes purely based on their density distribution in euclidean space without a restriction on the maximum number of clusters. We then trivially backtrack the cluster labels $l_i^{q'}$ from the votes $v_i^{q'}$ to the seeds $s_i$(see Fig. \ref{fig:clustering_viz})  to create an assignment $s_i^{q', L}$ 
where $L = l_i^{q'}$.
Following this, we also remove any spurious clusters with a member strength less than $0.05 \times M$ for rooms and  $0.01 \times M$ for walls, to finally create $C^R$ number of room clusters and $C^W$ number of wall clusters. To obtain the list of wall points belonging to a room, we perform an intersection operation on the room and wall point sets as described below:

\begin{equation}
    r^{k} = s^{R, k}
\end{equation}
 
\text{and}
   
\begin{equation}
    w^{m, k} =  r^{k} \cap s^{W, m } \text{  for  } k \in \mathbb{R}^{C^W}, m \in \mathbb{R}^{C^R},
\end{equation}
where $r^{k}$ is the set of points belonging to the $k^{th}$ room and $w^{m, k}$ is the set of points that belong to the $m^{th}$ wall of the $k^{th}$ room. Since not all walls belong to all the rooms, a large number of the intersections $w^{m, k}$ are a null set. For ease of notation, we ignore all such null sets and redefine  $w^{m, k}$ as $w^{m', k}$ where $m' \in \mathbb{R}^{C^{mk}}$ and $C^{mk}$ is the number of non-empty sets in $\{w^{m, k}\}_{m=1}^{C^W}$.

\begin{figure}
\centering
\includegraphics[width=\textwidth]{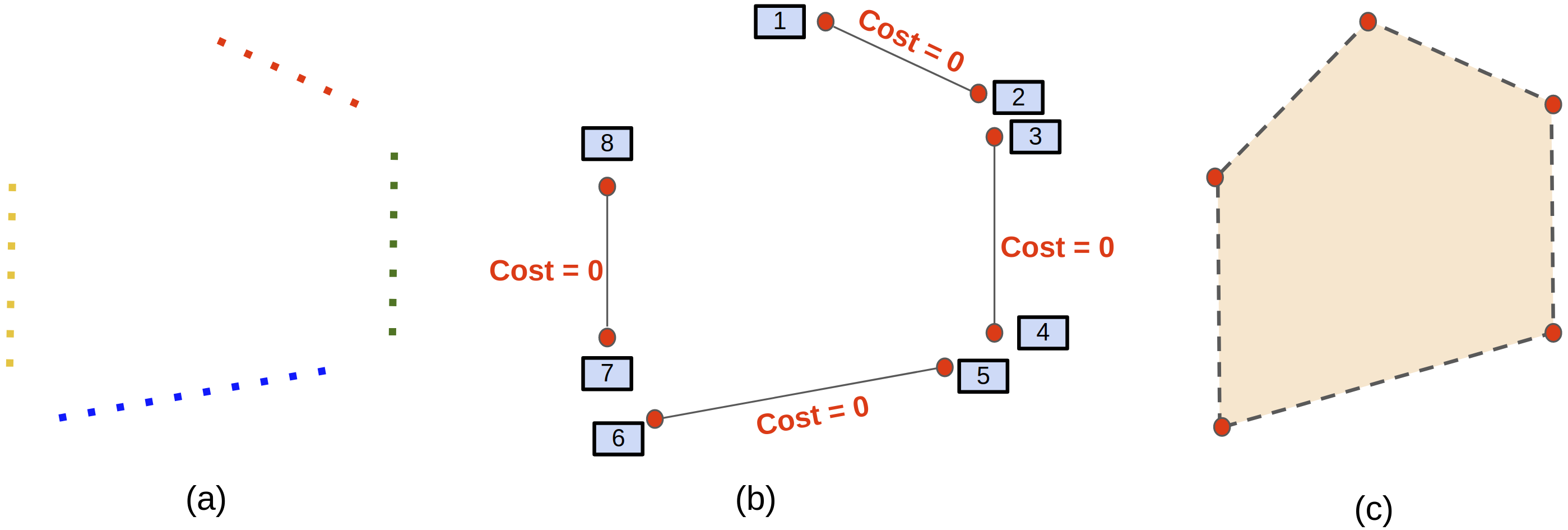}
\caption{\textbf{Room Perimeter Estimation} (a) - The set of clustered wall points which is the input to the perimeter estimation algorithm for a room. (b) - Ordering of the wall segment endpoints determined by the shortest path algorithm. (c) - Final room perimeter as a simple polygon generated by extruding the line segments to regress the polygon corners.}
\label{fig:perimeter_viz}
\end{figure}

\subsection{Room Perimeter Estimation}
\label{subsect:perimeter}

Given the inferred room points $r^k$, and the room wall points $\{w^{m', k}\}_{m'=1}^{C^{mk}}$ for a room with cluster label $k$,  we build upon the methods described in DeepPerimeter\cite{phalak2019deepperimeter} to delineate individual room perimeters. Since our final floorplan is 2D, we proceed by projecting all points in  $r^k$ and $\{w^{m', k}\}_{m'=1}^{C^{mk}}$ to the X-Y plane under the assumption that all walls are orthogonal to the X-Y plane. Let $w^{h, k}$ denote the set of points belonging to the $h^{th}$ wall of the $k^{th}$ room. By using RANSAC\cite{fischler1981random}, we a predict 2D line segment for all points in $w^{h, k}$ denoted by $p^{h, k}$, parameterized by the line segment endpoints. We remove any line segments in $\{p^{m', k}\}_{m'=1}^{C^{mk}}$ that are deemed to be duplicate, specifically if their the difference in their slope $\leq \theta_{min}$ and the difference in their bias $\leq \beta_{min}$. Any line segments with an angle $\leq \theta_{orth}$ with the nearest orthogonal axis are snapped to the align with the said axis. To allow for non-manhattan layouts while  also rectifying minor errors in the RANSAC line fitting we keep the value of $\theta_{orth}$ relatively low, as described in Sec. \ref{sect:experiments}.

In order to form a closed path along a set of nodes, as shown in Fig. \ref{fig:perimeter_viz}, we implement a modified version of the 2-opt algorithm from \cite{croes1958method} which provides a solution to the traveling-salesman problem. The set of nodes through which we wish to compute a shortest path is the set of start-points $\{p_1^{m', k}\}_{m'=1}^{C^{mk}}$ and end-points $\{p_2^{m', k}\}_{m'=1}^{C^{mk}}$ of the line segments. This effectively doubles the number of input nodes, but also provides a more optimal solution in cases of more complicated layout shapes as compared to \cite{phalak2019deepperimeter}, which use only the medians of $\{w^{m', k}\}_{m'=1}^{C^{mk}}$ as their  set of nodes. Since the pair of endpoints $p_1^{h, k}$ and $p_2^{h, k}$ of the segment $p^{h, k}$ should always be directly connected by an edge we set the cost of traversal for all such pairs of edges to $0$ in the optimization problem(see Fig. \ref{fig:perimeter_viz}).
\section{Datasets}
\label{sec:datasets}
In this section we describe the various datasets used for the training and evaluation of Scan2Plan. It is important to note that the system is able to achieve competitive performance on unseen real and synthetic datasets while being trained purely on a procedurally generated synthetic dataset.

\subsection{Synthetic Dataset}
\label{subsect:dataset_syn}
The key motivation behind training our networks on a fully synthetic dataset is rooted in the shortage of publicly available datasets with full 3D representations of indoor scenes with annotated floorplans. The synthetic dataset we generate is highly configurable in all aspects, which allows altering the distribution of samples in terms of the room shapes, sizes, configurations and noise patterns to match any target test dataset. Moreover the simplicity of the algorithm and the rapid speed of generation of such a dataset enables training on a large number of samples with a rich variety of possible layouts.

\begin{figure}
\centering
\includegraphics[width=0.8\textwidth]{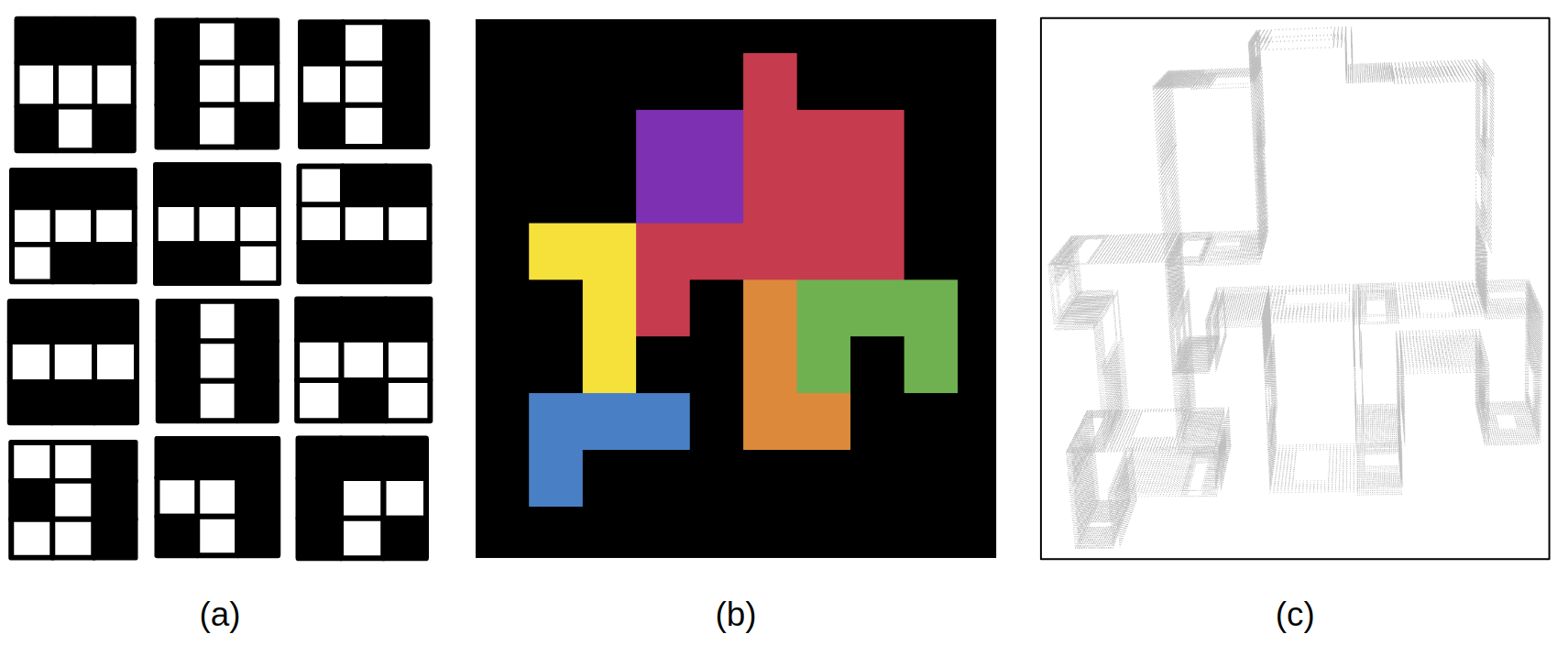}
\caption{ \textbf{Synthetic Dataset Generation}
(a) - Some of the shapes from the shape library. 
(b) - The labeled occupancy grid generated for a sample of the dataset.
(c) - The final point cloud generated from the occupancy grid, which can be used for training or testing the pipeline.}
\label{fig:synthetic_dataset}
\end{figure}

In order to generate this dataset, we start with a library of shapes shown in Fig. \ref{fig:synthetic_dataset}(a), which are simply various combinations of bits on a binary $3\times3$ kernel. To create a synthetic layout with $N_o$ rooms, we randomly select a shape from the library and place it on the center of a $32\times32$ grid, referred to as an occupancy grid, and assign a room label $l=0$ to it. To add the next $N_o - 1$ rooms and their labels, we iteratively select any of the adjacent unoccupied grid spaces and place another random shape from the library as long it does not overlap with the currently occupied spaces. Occasionally, we randomly create a ``super-room'', which is when we assign the same label $l$ to several connected shapes. An example can be seen in Fig. \ref{fig:synthetic_dataset}(b), where the red room is created from a combination of the library shapes, whereas the others are created from a single shape.

Once an occupancy grid is created with a random number of rooms from $0$ to $N_{omax}$, we iterate over the occupancy grid to identify wall locations and create a 3D representation of the wall plane by randomly sampling points on the plane assuming all the walls to have a fixed elevation, following which we are able to generate a 3D point cloud with two room labels and a wall label for each point. The two room labels are set to be identical if the point belongs to a single room, and they are distinct if the point belongs to two different rooms simultaneously. Each point is assumed to belong to a single wall only. To create even more variance in the room dimensions, we randomly scale the width and height of the rows and columns of the occupancy grid. We also cut out randomly sized rectangular blocks to represent missing points in scenarios where a door or window might be part of the wall. The resulting point cloud can be seen in Fig. \ref{fig:synthetic_dataset}(c). 

At the time of training we apply a random rotation and also a scaling for each of the X, Y and Z axes to each sample, and normalize the input to a $2m\times2m$ box in the first quadrant. The same normalization is enforced during inference as well. 

\subsection{Structured3D Dataset}

We use the Structured3D dataset\cite{zheng2019structured3d} to evaluate the quantitative and qualitative performance of our system. This dataset is a collection of 3500 apartment scans created by professional designers with a challenging variance in the room shapes and their configurations. Each sample in this dataset has upto 22 rooms, with an average of $5.79$ rooms per sample along with various annotations such as wall junctions, wall planes, room cuboids and scene semantics. An example layout can be seen in the bottom right of Fig. \ref{fig:results}.

\subsection{BKE Dataset}
\label{subsect:dataset_floorsp}
\begin{figure}
\centering
\includegraphics[width=\textwidth]{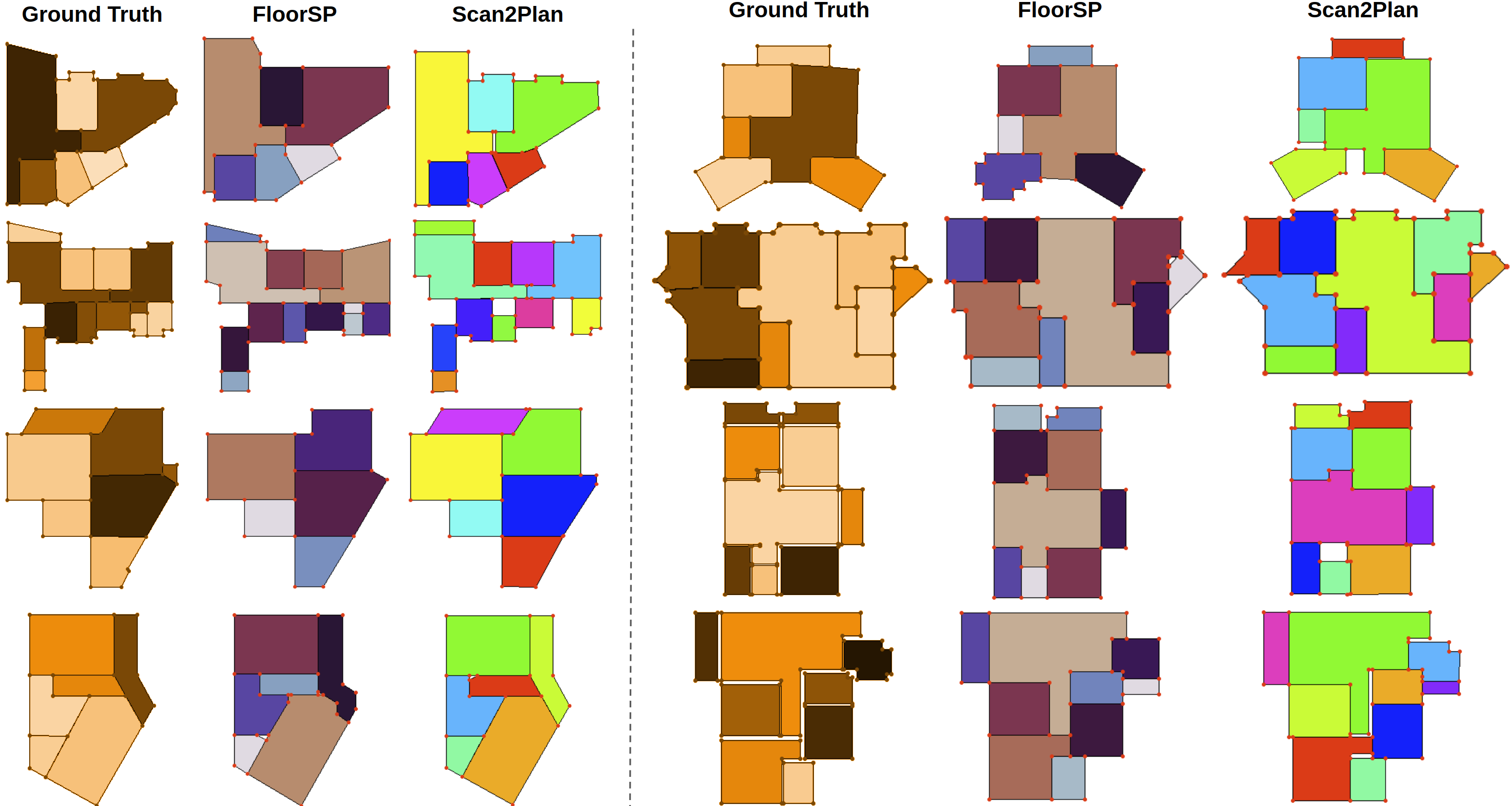}
\caption{\textbf{Results} Examples of floorplans predicted by FloorSP and Scan2Plan on samples from the Structured3D\cite{zheng2019structured3d} and BKE-$syn$\cite{BKE} datasets. From the results under the Scan2Plan column, the ability of our method to tackle complex non-Manhattan layouts accurately is clearly evident.}
\label{fig:results}
\end{figure}

We also use the BKE\cite{BKE} dataset from \cite{cjc2019floorsp} to evaluate the quantitative and qualitative performance of our system. Each sample in this dataset is captured using multiple LiDAR scans and has upto 13 rooms, with an average of $7.07$ rooms per sample.
Since we operate under the assumption that the input scan contains only the structural elements such as walls, doors and windows, we perform experiments by using two different versions of this dataset. In the first version which we shall refer to as BKE-$\it{syn}$, we construct a synthetic point cloud using the corner, edge and room annotations provided in the dataset. The samples from this dataset are clean, noise-free, and contain a uniform sampling of points along all walls. The second version, which we shall refer to as BKE-$\it{struct}$ is obtained by retaining points in the original scan that are nearer than $0.15m$ to the nearest corresponding point from the same scene in BKE-$\it{syn}$. Thus we obtain a subset of points of the original scan that represent the structural elements essential to floorplan estimation while discarding the internal clutter.
It is also possible to perform this filtering using publicly available semantic segmentation networks such as MinkowskiNet\cite{choy20194d} or ScanComplete\cite{dai2018scancomplete}. However, in our experiments, the pretrained MinkowskiNet performs poorly on unseen datasets and a corresponding problem-specific training dataset for BKE is unavailable. Hence, we rely instead on the annotations provided in the dataset itself to generate input point clouds for our method. An example of a layout from this dataset can be seen in the top-left of Fig. \ref{fig:results}.

\section{Experiments}
\label{sect:experiments}
We now detail the various experiments involved in generating a floorplan from a 3D scan and discuss the quantitative and qualitative results of our pipeline on the datasets discussed in Sec. \ref{sec:datasets}.

\subsection{Implementation Details}
All experiments are performed on a Nvidia GTX-Ti GPU and a Intel-XEON CPU with 8 cores. We set the number of points input to the network as $N=16384$ for training and evaluation by randomly subsampling the original point cloud. For the PointNet backbone in the voting module, we use the same architecture as in \cite{qi2019deep}. Since we normalize all the input points to lie in a $2m\times2m$ box, we set the radii of the four set abstraction layers to $[0.2, 0.4, 0.8, 1.2]$ respectively to enable computation of features at all possible scales. The number of feature channels is set to $C=256$ for every seed point, the number of seed points to be created are $M=1024$, and the loss balancing factor is set to $\alpha=10$. The first vote generator(for rooms) in Fig. \ref{fig:votenet_architecture} has layers with output of sizes of $[256, 256, 6]$, whereas the second vote generator(for walls) has layers with output sizes of $[256, 256, 3]$.

For the DBSCAN clustering step, we set the value of $\epsilon=5cm$ for room votes and $\epsilon=2.5cm$ for wall votes. $\epsilon$ is in some sense a measure of the maximum distance between two points for the algorithm to group them into the same cluster. During the perimeter estimation step, we set $\theta_{min} = 15$\textdegree, $\beta_{min}=15cm$ and $\theta_{orth}=15$\textdegree. The synthetic training dataset in Sec. \ref{subsect:dataset_syn} is generated with a maximum of just $N_{omax} = 10$ rooms per sample, however during evaluation, the network is able to achieve excellent performance when the number of rooms is greater than $N_{omax}$ as well.

\begin{table}
\caption {\textbf{Quantitative Results of Scan2Plan} compared against state-of-the-art floorplan estimation methods. We follow the metrics described in FloorSP\cite{cjc2019floorsp}. Our method outperforms the state-of-the art in Precision and Recall metrics(higher is better) and Runtime(lower is better) despite being trained on a fully synthetic dataset.}
    \resizebox{\textwidth}{!}{
    \begin{tabular}{|c|>{\centering\arraybackslash}p{0.15\textwidth}|>{\centering\arraybackslash}p{0.15\textwidth}|>{\centering\arraybackslash}p{0.15\textwidth}|>{\centering\arraybackslash}p{0.15\textwidth}|>{\centering\arraybackslash}p{0.15\textwidth}|>{\centering\arraybackslash}p{0.15\textwidth}|}
    \hline
      \textbf{Dataset(Method)} & \textbf{Corner Precision} & \textbf{Corner Recall} & \textbf{Edge Precision} & \textbf{Edge Recall} & \textbf{Room Precision} & \textbf{Room Recall}\\

      \hline
      Structured3D(FloorSP) & 0.838 & 0.804 & 0.778 & 0.753  & 0.795 & 0.866\\
      \hline
      Structured3D(Ours) & \textbf{0.856} & \textbf{0.834} & \textbf{0.813} & \textbf{0.795}  & \textbf{0.880} & \textbf{0.887}\\
      \hline
      \hline
      
      
      BKE-$syn$(FloorSP) & 0.926 & 0.833 & 0.830 & 0.750  & 0.947 & \textbf{0.944}\\
      \hline
      BKE-$syn$(Ours) & \textbf{0.959} & \textbf{0.875} & \textbf{0.902} & \textbf{0.823}  & \textbf{0.962} & 0.915\\
      \hline
      \hline
      
      BKE-$struct$(FloorSP) & 0.830 & \textbf{0.659} & 0.697 & \textbf{0.557}  & 0.808 & \textbf{0.690}\\
      \hline
      BKE-$struct$(Ours) & \textbf{0.842} & 0.607 & \textbf{0.714} & 0.518  & \textbf{0.870} & 0.610\\
      \hline
    \end{tabular}}
    \vspace{0.25cm}
    \newline
    \resizebox{0.4\textwidth}{!}{
    \begin{tabular}{|>{\centering\arraybackslash}p{0.275\textwidth}|>{\centering\arraybackslash}p{0.195\textwidth}|}
        \hline
        \textbf{Method} &  \textbf{Runtime(s)}\\
        \hline
        FloorSP & 487.50 \\
        \hline        
        Scan2Plan & \textbf{3.96} \\
    \hline
    \end{tabular}}
\label{tab:metrics}
\end{table}

\subsection{Results and Metrics}
In order to compare our results to the state-of-the-art methods, we report on the metrics introduced by FloorSP\cite{cjc2019floorsp} and generate precision and recall values for all the corner locations, edges and IOU(Intersection-over-Union) values for rooms. It should be noted that we do not parameterize the layout as a joint global graph but instead consider each room as an independent simple polygon for computing metrics. Similar to the FloorSP approach, we jointly transform and project both the ground truth and prediction corners and edges onto a $256\times256$ image grid and use the following rules for calculating metrics :

\textbf{Corners: }
The list of corners is a concatenation of all room corners irrespective of their location. This implies that even though multiple corners might have the same 2D coordinates, we do not unify them if they belong to distinct rooms. Following this logic for both ground truth and predicted corners, we use a Hungarian matrix to solve the assignment problem and compute precision and recall wherein a prediction is considered true-positive if there exists a unique corner in the GT within a distance of 10 pixels from it.

\textbf{Edges: }
Similar to the corners, we concatenate edges across all rooms and consider an edge to be a true-positive if both its corners are true-positives

\textbf{Room: } 
A room is considered to be a true-positive if it has an IOU score of over 0.7 with a unique room from the ground truth. In contrast to FloorSP, we already resolve any room overlaps in post-processing, so any room polygons generated by Scan2Plan are guaranteed to be mutually exclusive in the 2D space.

Table \ref{tab:metrics} shows the quantitative performance(accuracy and end-to-end runtime) of our method and also compares it with other state-of-the-art methods for all the datasets under consideration.

\begin{figure}
\centering
\includegraphics[width=\textwidth]{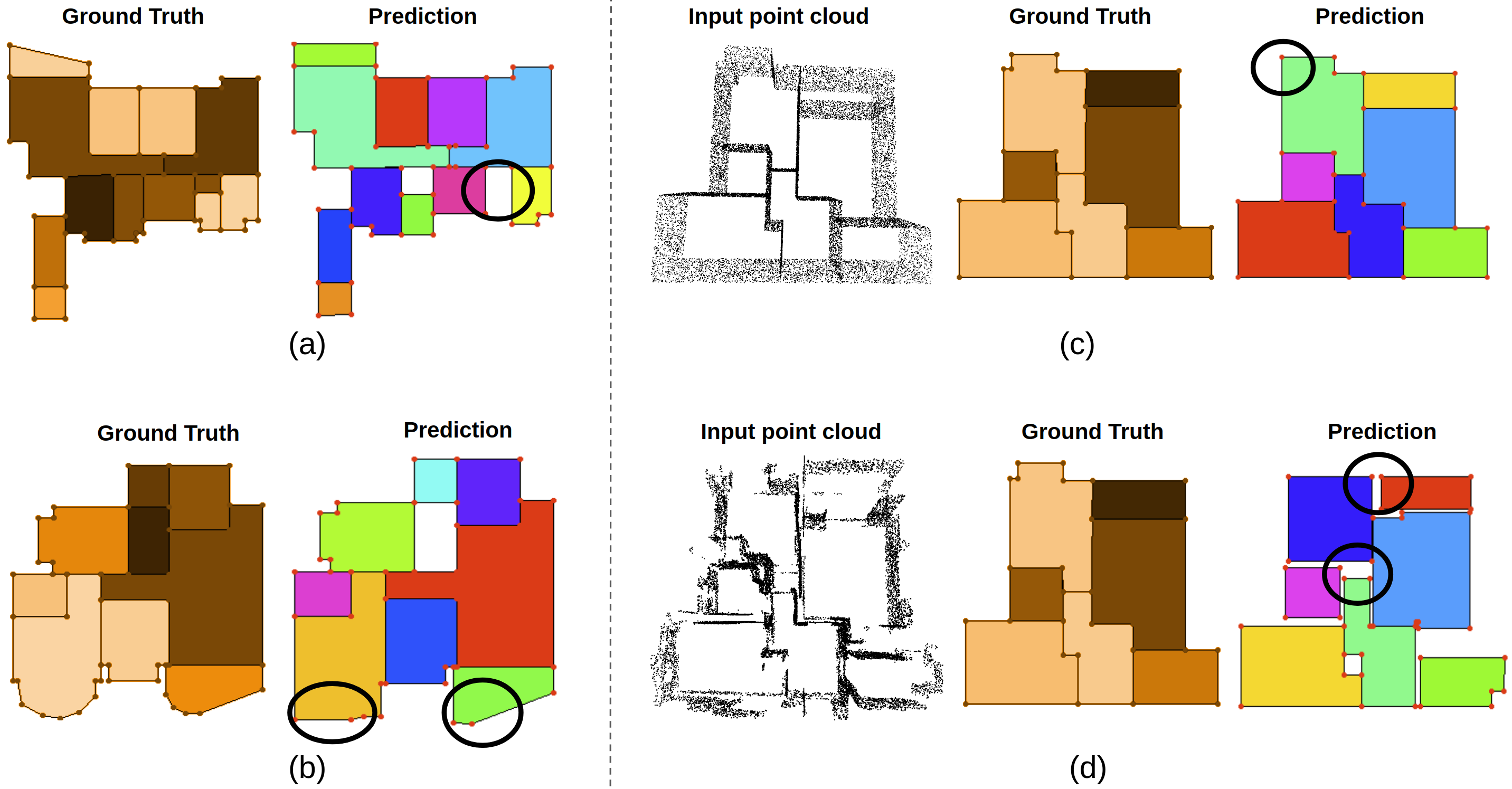}
\caption{\textbf{Failure Cases} (a) - The effects of erroneous clustering on very cluttered layouts where the smaller rooms are absent in the prediction. (b) - Inaccurate room perimeter estimation in cases with curved walls. (c) -  An adverse effect of the wall culling described in Sec. \ref{subsubsect:clustering}. (d) - Shortcomings of using a non-global method on inputs with missing points results in a non-compact floorplan.}
\label{fig:failure}
\end{figure}

\subsection{Discussion}

Scan2Plan is able to generate accurate floorplans for a variety of shapes as shown in Fig. \ref{fig:results}. It can be seen that even though we train the voting network with only manhattan style layouts, the room and wall clustering is equally successful on complex non-manhattan style layouts, and also room shapes that are not present in the training set. This is possible due the variety we introduce in the dataset, importantly the randomized rotations we apply on the input, which trains the network on 
rooms and walls that are not axis-aligned.

In case of extremely cluttered layouts such as ones in  Fig. \ref{fig:failure}(a), some small rooms are omitted because of a combination of firstly - imperfect clustering, where two small connected rooms are assigned the same label and secondly - a post processing step in the perimeter estimation where we omit walls with a low number of points as described in Sec. \ref{subsubsect:clustering}. One more effect of this wall culling can be seen in Fig. \ref{fig:failure}(c) where a small notch in the floorplan is omitted. Another limitation of our method is exposed via scans with curved walls such as Fig. \ref{fig:failure}(b), where in the ground truth, the curvature is expressed by a series of corners, whereas the clustering module combines multiple of such very small segments into one large segment.

On the Sructured3D dataset(see Table \ref{tab:metrics}), we consistently outperform the state-of-the-art in all categories. On the BKE-$syn$, FloorSP has a higher room recall due to the fact that our method is susceptible to grouping multiple rooms together, or omitting a room altogether in scenarios discussed in Fig. \ref{fig:failure}. On the BKE-$struct$ dataset, our method achieves a higher precision, but a lower recall value than the state-of-the-art for all categories. This implies that our method tends to selectively generate corner, edge and room primitives that are more accurate compared to FloorSP, which generates a larger number of less accurate primitives. A result from BKE-$struct$ can be observed in Fig. \ref{fig:failure}(d), where the input scans themselves are missing multiple sections of the structural elements in the scene due to the imperfect filtering procedure described in Sec. \ref{subsect:dataset_floorsp}.


In future iterations of this work, there exist multiple areas of this pipeline upon which we aim to improve. The perimeter estimation stage can be combined with the clustering stage to form an end-to-end trainable system. Such a system would take the same point cloud as an input and provided either a rasterized version of the layout or a series of polygons for each room in the scene. This would make the system even more computationally efficient, robust to additive and subtractive noise in the inputs and also assist the clustering task with the error that is back-propogated from the perimeter estimation task. In the case of multi-level apartments and buildings, an additional cluster label to identify the floor-level can be added and a complete building blueprint can be generated by stacking such individual layouts. Furthermore, it should also be possible for the backbone network to generate features on the 3D scan to perform an additional task of identifying door and window locations to further add another dimension of semantic information to the current inference.

\section{Conclusion}
\label{sect:conclusion}

The method Scan2Plan proposed in this paper is an efficient and robust approach for generating accurate floorplans from 3D scans of indoor scenes. Scan2Plan relies on a deep network to perform room and wall clustering and is fully trainable purely on a synthetic dataset described in this paper. The latter part of the pipeline which predicts individual room perimeters using procedural algorithms, is highly parallelizable and as a whole, the method outperforms current state-of-the art techniques in both speed and accuracy. Scan2Plan is able to generate layouts of scenes without assumptions regarding the shape, size, number and configuration of rooms which renders it valuable for floorplan estimation from 3D data in the wild.

\newpage

\bibliographystyle{splncs04}
\bibliography{main}

\begin{thebibliography}{10}
\providecommand{\url}[1]{\texttt{#1}}
\providecommand{\urlprefix}{URL }
\providecommand{\doi}[1]{https://doi.org/#1}

\bibitem{BKE}
\url{http://www.ke.com/}

\bibitem{avetisyan2019scan2cad}
Avetisyan, A., Dahnert, M., Dai, A., Savva, M., Chang, A.X., Nie{\ss}ner, M.:
  Scan2cad: Learning cad model alignment in rgb-d scans. In: Proceedings of the
  IEEE Conference on Computer Vision and Pattern Recognition. pp. 2614--2623
  (2019)

\bibitem{ayad2007cumulative}
Ayad, H.G., Kamel, M.S.: Cumulative voting consensus method for partitions with
  variable number of clusters. IEEE transactions on pattern analysis and
  machine intelligence  \textbf{30}(1),  160--173 (2007)

\bibitem{caron2018deep}
Caron, M., Bojanowski, P., Joulin, A., Douze, M.: Deep clustering for
  unsupervised learning of visual features. In: Proceedings of the European
  Conference on Computer Vision (ECCV). pp. 132--149 (2018)

\bibitem{choy20194d}
Choy, C., Gwak, J., Savarese, S.: 4d spatio-temporal convnets: Minkowski
  convolutional neural networks. In: Proceedings of the IEEE Conference on
  Computer Vision and Pattern Recognition. pp. 3075--3084 (2019)

\bibitem{croes1958method}
Croes, G.A.: A method for solving traveling-salesman problems. Operations
  research  \textbf{6}(6),  791--812 (1958)

\bibitem{dai2018scancomplete}
Dai, A., Ritchie, D., Bokeloh, M., Reed, S., Sturm, J., Nie{\ss}ner, M.:
  Scancomplete: Large-scale scene completion and semantic segmentation for 3d
  scans. In: Proceedings of the IEEE Conference on Computer Vision and Pattern
  Recognition. pp. 4578--4587 (2018)

\bibitem{dasgupta2016delay}
Dasgupta, S., Fang, K., Chen, K., Savarese, S.: Delay: Robust spatial layout
  estimation for cluttered indoor scenes. In: Proceedings of the IEEE
  conference on computer vision and pattern recognition. pp. 616--624 (2016)

\bibitem{dimitriadou2001voting}
Dimitriadou, E., Weingessel, A., Hornik, K.: Voting-merging: An ensemble method
  for clustering. In: International Conference on Artificial Neural Networks.
  pp. 217--224. Springer (2001)

\bibitem{ester1996density}
Ester, M., Kriegel, H.P., Sander, J., Xu, X., et~al.: A density-based algorithm
  for discovering clusters in large spatial databases with noise. In: Kdd.
  vol.~96, pp. 226--231 (1996)

\bibitem{fischler1981random}
Fischler, M.A., Bolles, R.C.: Random sample consensus: a paradigm for model
  fitting with applications to image analysis and automated cartography.
  Communications of the ACM  \textbf{24}(6),  381--395 (1981)

\bibitem{he2017mask}
He, K., Gkioxari, G., Doll{\'a}r, P., Girshick, R.: Mask r-cnn. In: Proceedings
  of the IEEE international conference on computer vision. pp. 2961--2969
  (2017)

\bibitem{He_2017_ICCV}
He, K., Gkioxari, G., Dollar, P., Girshick, R.: Mask r-cnn. In: The IEEE
  International Conference on Computer Vision (ICCV) (Oct 2017)

\bibitem{hsiao2019flat2layout}
Hsiao, C.W., Sun, C., Sun, M., Chen, H.T.: Flat2layout: Flat representation for
  estimating layout of general room types. arXiv preprint arXiv:1905.12571
  (2019)

\bibitem{iqbal2012semi}
Iqbal, A.M., Moh'd, A., Khan, Z.: Semi-supervised clustering ensemble by
  voting. arXiv preprint arXiv:1208.4138  (2012)

\bibitem{cjc2019floorsp}
Jiacheng~Chen, Chen~Liu, J.W.Y.F.: Floor-sp: Inverse cad for floorplans by
  sequential room-wise shortest path. In: The IEEE International Conference on
  Computer Vision (ICCV) (2019)

\bibitem{kruzhilov2019double}
Kruzhilov, I., Romanov, M., Konushin, A.: Double refinement network for room
  layout estimation  (2019)

\bibitem{Lee_2017_ICCV}
Lee, C.Y., Badrinarayanan, V., Malisiewicz, T., Rabinovich, A.: Roomnet:
  End-to-end room layout estimation. In: The IEEE International Conference on
  Computer Vision (ICCV) (Oct 2017)

\bibitem{li2018pointcnn}
Li, Y., Bu, R., Sun, M., Wu, W., Di, X., Chen, B.: Pointcnn: Convolution on
  x-transformed points. In: Advances in neural information processing systems.
  pp. 820--830 (2018)

\bibitem{liu2018floornet}
Liu, C., Wu, J., Furukawa, Y.: Floornet: A unified framework for floorplan
  reconstruction from 3d scans. In: Proceedings of the European Conference on
  Computer Vision (ECCV). pp. 201--217 (2018)

\bibitem{liu2016ssd}
Liu, W., Anguelov, D., Erhan, D., Szegedy, C., Reed, S., Fu, C.Y., Berg, A.C.:
  Ssd: Single shot multibox detector. In: European conference on computer
  vision. pp. 21--37. Springer (2016)

\bibitem{murali2017indoor}
Murali, S., Speciale, P., Oswald, M.R., Pollefeys, M.: Indoor scan2bim:
  Building information models of house interiors. In: 2017 IEEE/RSJ
  International Conference on Intelligent Robots and Systems (IROS). pp.
  6126--6133. IEEE (2017)

\bibitem{phalak2019deepperimeter}
Phalak, A., Chen, Z., Yi, D., Gupta, K., Badrinarayanan, V., Rabinovich, A.:
  Deepperimeter: Indoor boundary estimation from posed monocular sequences.
  arXiv preprint arXiv:1904.11595  (2019)

\bibitem{phan2018dgcnn}
Phan, A.V., Le~Nguyen, M., Nguyen, Y.L.H., Bui, L.T.: Dgcnn: A convolutional
  neural network over large-scale labeled graphs. Neural Networks
  \textbf{108},  533--543 (2018)

\bibitem{qi2019deep}
Qi, C.R., Litany, O., He, K., Guibas, L.J.: Deep hough voting for 3d object
  detection in point clouds. In: Proceedings of the IEEE International
  Conference on Computer Vision. pp. 9277--9286 (2019)

\bibitem{qi2017pointnet}
Qi, C.R., Su, H., Mo, K., Guibas, L.J.: Pointnet: Deep learning on point sets
  for 3d classification and segmentation. In: Proceedings of the IEEE
  conference on computer vision and pattern recognition. pp. 652--660 (2017)

\bibitem{qi2017pointnet++}
Qi, C.R., Yi, L., Su, H., Guibas, L.J.: Pointnet++: Deep hierarchical feature
  learning on point sets in a metric space. In: Advances in neural information
  processing systems. pp. 5099--5108 (2017)

\bibitem{qi20173d}
Qi, X., Liao, R., Jia, J., Fidler, S., Urtasun, R.: 3d graph neural networks
  for rgbd semantic segmentation. In: Proceedings of the IEEE International
  Conference on Computer Vision. pp. 5199--5208 (2017)

\bibitem{redmon2016you}
Redmon, J., Divvala, S., Girshick, R., Farhadi, A.: You only look once:
  Unified, real-time object detection. In: Proceedings of the IEEE conference
  on computer vision and pattern recognition. pp. 779--788 (2016)

\bibitem{ren2015faster}
Ren, S., He, K., Girshick, R., Sun, J.: Faster r-cnn: Towards real-time object
  detection with region proposal networks. In: Advances in neural information
  processing systems. pp. 91--99 (2015)

\bibitem{schubert2017dbscan}
Schubert, E., Sander, J., Ester, M., Kriegel, H.P., Xu, X.: Dbscan revisited,
  revisited: why and how you should (still) use dbscan. ACM Transactions on
  Database Systems (TODS)  \textbf{42}(3),  1--21 (2017)

\bibitem{shen2018mining}
Shen, Y., Feng, C., Yang, Y., Tian, D.: Mining point cloud local structures by
  kernel correlation and graph pooling. In: Proceedings of the IEEE conference
  on computer vision and pattern recognition. pp. 4548--4557 (2018)

\bibitem{shukla2018semi}
Shukla, A., Cheema, G.S., Anand, S.: Semi-supervised clustering with neural
  networks. arXiv preprint arXiv:1806.01547  (2018)

\bibitem{tatarchenko2017octree}
Tatarchenko, M., Dosovitskiy, A., Brox, T.: Octree generating networks:
  Efficient convolutional architectures for high-resolution 3d outputs. In:
  Proceedings of the IEEE International Conference on Computer Vision. pp.
  2088--2096 (2017)

\bibitem{wang2017cnn}
Wang, P.S., Liu, Y., Guo, Y.X., Sun, C.Y., Tong, X.: O-cnn: Octree-based
  convolutional neural networks for 3d shape analysis. ACM Transactions on
  Graphics (TOG)  \textbf{36}(4),  1--11 (2017)

\bibitem{wang2019dynamic}
Wang, Y., Sun, Y., Liu, Z., Sarma, S.E., Bronstein, M.M., Solomon, J.M.:
  Dynamic graph cnn for learning on point clouds. ACM Transactions on Graphics
  (TOG)  \textbf{38}(5),  1--12 (2019)

\bibitem{xie2016unsupervised}
Xie, J., Girshick, R., Farhadi, A.: Unsupervised deep embedding for clustering
  analysis. In: International conference on machine learning. pp. 478--487
  (2016)

\bibitem{zhang2013estimating}
Zhang, J., Kan, C., Schwing, A.G., Urtasun, R.: Estimating the 3d layout of
  indoor scenes and its clutter from depth sensors. In: Proceedings of the IEEE
  International Conference on Computer Vision. pp. 1273--1280 (2013)

\bibitem{zhang2019edge}
Zhang, W., Zhang, W., Gu, J.: Edge-semantic learning strategy for layout
  estimation in indoor environment. IEEE transactions on cybernetics  (2019)

\bibitem{zhao2017pyramid}
Zhao, H., Shi, J., Qi, X., Wang, X., Jia, J.: Pyramid scene parsing network.
  In: Proceedings of the IEEE conference on computer vision and pattern
  recognition. pp. 2881--2890 (2017)

\bibitem{zheng2019structured3d}
Zheng, J., Zhang, J., Li, J., Tang, R., Gao, S., Zhou, Z.: Structured3d: A
  large photo-realistic dataset for structured 3d modeling. arXiv preprint
  arXiv:1908.00222  (2019)

\bibitem{zhou2018voxelnet}
Zhou, Y., Tuzel, O.: Voxelnet: End-to-end learning for point cloud based 3d
  object detection. In: Proceedings of the IEEE Conference on Computer Vision
  and Pattern Recognition. pp. 4490--4499 (2018)

\bibitem{zou2018layoutnet}
Zou, C., Colburn, A., Shan, Q., Hoiem, D.: Layoutnet: Reconstructing the 3d
  room layout from a single rgb image. In: Proceedings of the IEEE Conference
  on Computer Vision and Pattern Recognition. pp. 2051--2059 (2018)

\end{thebibliography}
\end{document}